\documentclass{article}

\usepackage{authblk}
\usepackage{arxiv}
\usepackage{float}

\usepackage[utf8]{inputenc} 
\usepackage[T1]{fontenc}    
\usepackage{hyperref}       
\usepackage{url}            
\usepackage{booktabs}       
\usepackage{amsfonts}       
\usepackage{nicefrac}       
\usepackage{microtype}      

\usepackage{graphicx}
\usepackage[numbers]{natbib}
\usepackage{doi}

\usepackage{listings}
\usepackage{tikz}
\usetikzlibrary{arrows.meta,positioning,shapes.geometric,fit,calc}

\tikzset{
  base/.style={rounded corners, draw, thick, fill=blue!7, inner sep=2pt, align=left},
  exc/.style={rounded corners, draw, thick, fill=orange!10, inner sep=6pt, align=left},
  cond/.style={draw, thick, rectangle, rounded corners, fill=gray!8, inner sep=6pt, align=left},
  labelnode/.style={draw, thick, ellipse, fill=green!10, inner sep=4pt},
  tinylabel/.style={font=\footnotesize\ttfamily},
  arr/.style={-Latex, thick},
  dashedarr/.style={-Latex, thick, dashed},
}

\title{Comparative Analysis of FOLD-SE vs. FOLD-R++ in Binary Classification and XGBoost in Multi-Category Classification}

\author[1,4]{Akshay Murthy}
\author[2,4]{Shawn Sebastian}
\author[2,4]{Manil Shangle}
\author[3,4]{Huaduo Wang}
\author[3,4]{Sopam Dasgupta}
\author[3,4]{Gopal Gupta}
\affil[1]{Liberty High School, Frisco, Texas}
\affil[2]{Lebanon Trail High School, Frisco, Texas}
\affil[3]{Department of Computer Science, University of Texas at Dallas, TX}
\affil[4]{Applied Logic, Programming Languages, and Systems (ALPS) Lab, University of Texas at Dallas, TX}

\renewcommand{\shorttitle}{\textit{arXiv} Template}

\hypersetup{
pdftitle={A template for the arxiv style},
pdfsubject={q-bio.NC, q-bio.QM},
pdfauthor={David S.~Hippocampus, Elias D.~Striatum},
pdfkeywords={Explainable AI, Rule-based Learning, FOLD-SE, XGBoost, Machine Learning Interpretability},
}

\begin{document}
\maketitle

\begin{abstract}
	Recently, the demand for Machine Learning (ML) models that can effectively balance accuracy, efficiency, and interpreability has grown significantly. Traditionally, there has been a tradeoff between accuracy and explainability in predictive models, with models such as Neural Networks achieving high accuracy on complex datasets while sacrificing internal transparency. As such, new rule-based algorithms such as FOLD-SE have been developed that provide tangible justification for predictions in the form of interpretable rule sets. The primary objective of this study was to compare FOLD-SE and FOLD-R++, both rule-based classifiers, in binary classification and evaluate how FOLD-SE performs against XGBoost, a widely used ensemble classifier, when applied to multi-category classification. We hypothesized that because FOLD-SE can generate a condensed rule set in a more explainable manner, it would lose upwards of an average of 3 percent in accuracy and F1 score when compared with XGBoost and FOLD-R++ in multiclass and binary classification, respectively. The research used data collections for classification, with accuracy, F1 scores, and processing time as the primary performance measures. Outcomes show that FOLD-SE is superior to FOLD-R++ in terms of binary classification by offering fewer rules but losing a minor percentage of accuracy and efficiency in processing time; in tasks that involve multi-category classifications, FOLD-SE is more precise and far more efficient compared to XGBoost, in addition to generating a comprehensible rule set. The results point out that FOLD-SE is a better choice for both binary tasks and classifications with multiple categories. Therefore, these results demonstrate that rule-based approaches like FOLD-SE can bridge the gap between explainability and performance, highlighting their potential as viable alternatives to black-box models in diverse classification tasks.

\end{abstract}


\section{Introduction}
The development of accurate Machine Learning (ML) systems to make human-like decisions has been a topic of interest for many in the past decades. As the limits of model accuracy are pushed, explainability in AI models has received increased attention as large-scale models tend to have corporate applications, often influencing the decisions of the company and individuals. Many state-of-the-art models, such as XG-Boost and Multilayer Perceptron, have very high prediction accuracy. Due to this, these  models are used in many real-world applications, such as informing long-term investment decisions in finance, achieving high-quality performance \cite{zolotareva2021xgboost}. However, their underlying logic is produced by convoluted and complex mathematical processes. Thus, in this sense, many current AI classifiers are regarded as “black boxes,” providing no tangible justification for their predictions and leaving individuals unclear on how the model came to its result \cite{umdearborn2021blackbox}.

\begin{figure}[!htbp]
    \centering
    \includegraphics[width=0.5\linewidth]{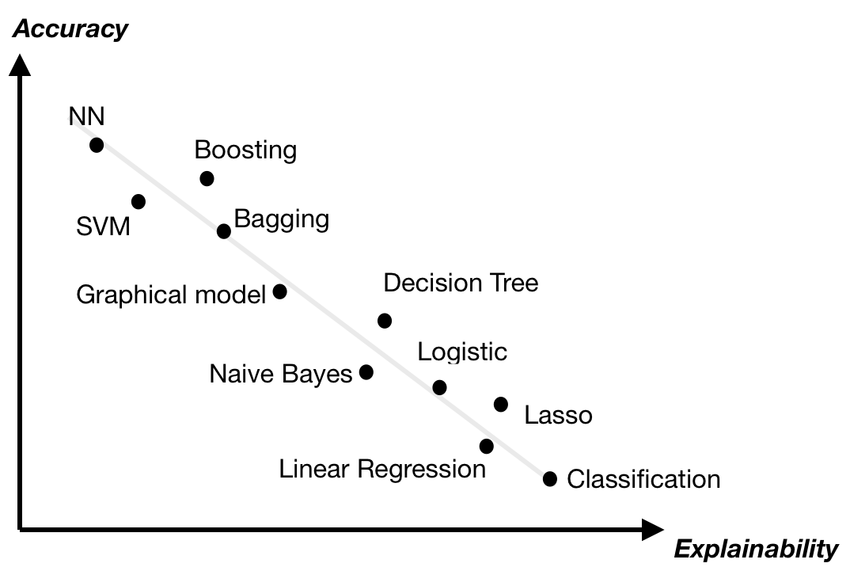}
    \caption{We can see the general trend of many state-of-the-art ML models: as accuracy increases, explainability conversely decreases (adapted from \cite{researchgate2019xai}).
}
    \label{fig:placeholder}
\end{figure}

Explainability in ML applications is highly important, as many corporate and medical decisions are dependent on the results of a trained ML model. Suppose those affected by the decision ask for a rationale or justification for the decision. In that case, if the ML model is not explainable, the business will face difficulty when justifying its decisions. This could result in ethical and legal issues, especially in controversial conversations. A typical example arises in the bank loan problem, where banks use an ML model to determine whether a customer should be given a loan or not, and the model's parameters are information about the customer. Oftentimes, if a loan is rejected, then the respondent would naturally question it, to which the business would struggle to respond, as conveying the weights and biases of a model is not sufficient. This can be especially worrisome if the model consistently is biased and makes judgments against or in favor of a particular group of individuals \cite{hale2022aibias}. Therefore, it would be helpful for an ML model to justify its decisions, preferably in a tangible way that humans can understand, so the reasoning behind decisions can be best relayed. Explainable AI (XAI) would serve this purpose, as XAI techniques offer tangible justification and comprehensibility for the end user.

In the case of these models, explainability inversely decreases as accuracy improves, as more and more complex model parameters are created to allow for specific pattern recognition and intricate generalizations. This accuracy-explainability tradeoff is a critical metric to evaluate model applicability and predictive power pragmatically, and will be a key point of focus in this paper. Our primary goal in this paper is to assess the potential of increasing both accuracy and explainability with a single model, namely FOLD-SE. We also hope to compare its performance against its predecessor, FOLD-R++.

Recently, the Applied Logic and Programming Systems (ALPS) Lab at the University of Texas at Dallas, led by Dr. Gopal Gupta, implemented s(CASP), a programming language that finds solutions to answer set queries expressed as rules and predicates, into an ML model, creating FOLD-R++. The model produced an easily comprehensible rule set, which was used to generate predictions. Although this model was limited to binary classification, it achieved comparable results in terms of accuracy and F1 score to other state-of-the-art classifiers while also incorporating explainability \cite{wang2021foldrpp}. 

More recently, FOLD-SE was introduced by ALPS for multiclass classification tasks, which expanded on the previous FOLD-R++, which was restricted to binary tasks. This model, built on s(CASP), uses the same rule set approach as FOLD-R++, except that it generates more simplified and comprehensible rule sets for justification, allowing the model to achieve heightened levels of explainability \cite{foldse2022}. In this paper, we will be evaluating FOLD-SE for both binary and multicategorical classification to explore accuracy and explainability.

As this model seemingly makes significant progress concerning the accuracy-explainability trade-off, we questioned whether FOLD-SE’s increased accuracy could coexist with increasing explainability through condensed rule sets, potentially breaking the inverse accuracy-explainability trend that exists in model analysis. We hypothesized that there would consequently be a slight decrease of \~3\% in accuracy as the model may sacrifice some predictive power to be more explainable, a trend that has been largely consistent with state-of-the-art models. To evaluate this, we tested FOLD-SE in comparison to FOLD-R++ for binary classification and XG-Boost for multi-categorical classification. We found that in contrast to FOLD-R++ for binary classification, FOLD-SE generated significantly fewer rules, highly improving its model interpretability (Table 1), and had almost the same average accuracy for all eight datasets we tested on, but was much less efficient in terms of processing time, taking more time in seven out of eight datasets compared to FOLD-R++. In comparison to XG-Boost for multi-class classification, FOLD-SE had a significantly higher average accuracy and efficiency in processing time for all seven datasets we tested on, even achieving complete accuracy on two datasets that XG-Boost could not. As FOLD-SE has very high model interpretability and explainability, the findings of FOLD-SE’s superior accuracy rates are groundbreaking, as it is an example of both accuracy and explainability directly increasing. Although disproving our initial hypothesis, these results suggest how models can be created using clever techniques, such as rule-based logic systems, to maximize both explainability and accuracy, creating new standards of model performance and potentially new real-world applications for models with explainability.

\subsection{Related Work}
\label{sec:headings}
The area of Explainable AI (XAI) is quickly changing, with new improvements showing its significance in different fields. For example, in healthcare, using methods like SHAP (Shapley Additive exPlanations) and LIME (Local Interpretable Model-agnostic Explanations) has become very useful for predicting complex results such as myocardial infarction and skin cancer \cite{saranya2023xai}. These methods provide clearer understandings of model predictions, increasing trust in AI and improving decisions in clinical environments. Additionally, in the field of finance, XAI methods such as the TREPAN decision tree are used to enhance transparency in predictions of credit card default \cite{saranya2023xai}. Such techniques allow users to understand on which basis neural networks make decisions and generate outputs, transforming the neural network into a understandable decision tree. Ongoing research in new XAI techniques shows the cruciality of the field for making machine learning models easier to interpret, especially in high-risk situations where understanding the rationale behind a decision is equally important as the decision itself.

\section{Materials and Methodology}
In this experiment, we had two objectives:
\begin{enumerate}
    \item Analyze the differences between FOLD-SE and FOLD-R++ in terms of scalability and accuracy in binary classification
    \item Analyze the performance of FOLD-SE against XG Boost in multicategory classification.
\end{enumerate}

We were able to implement FOLD-R++ programatically in Python through its open-source availability on Github \cite{hwd4042021foldrpp}. We accessed FOLD-SE online through a web API\cite{foldse2024tool}. This web API returned all of our performance metrics, including F1 scores, elapsed time, and all rules generated. We i

In our experiment, we used eight binary classification datasets to analyze the difference between FOLD-SE and FOLD-R++. Specifically, we tested on the datasets available in the FOLD-R++ GitHub repositories list of sample datasets \cite{hwd4042021foldrpp}, including the CSV files for kidney, autism, ecoli, heart, cars, sonar, mushroom, and adult. 

We ran Fold-R++ locally on a MacBook Air M2 chip, using Visual Studio Code version 1.92. We ran one data set at a time, collecting the accuracy, F1 score, base rules, exception rules, and elapsed processing time for each. We simultaneously utilized FOLD-SE on the same datasets using the online FOLD-SE web tool, keeping all the default hyperparameters set, such as the train-test ratio, level of exceptions, and tail ratio. We also ran one dataset at a time, collecting the same parameters as we did for FOLD-R++. F1 score and number of base and exception rules were all analyzed to assess accuracy and explainability in each model.

Once the data was collected, the raw percentage accuracy and F1 score were compared between the two applications to analyze precision. F1 score is defined as $$\frac{2\cdot \mathrm{Precision} \cdot \mathrm{Recall}}{\mathrm{Precision}+\mathrm{Recall}}$$ where $$\mathrm{Precision} = \frac{TP}{TP+FP}$$and $$\mathrm{Recall}= \frac{TP}{TP+FN}$$

\begin{description}
    \item[] $TP$ is number of true positives
    \item[] $FP$ is number of false positives
    \item[] $FN$ is number of false negatives
\end{description}
Additionally, the number of base rules and exception rules was compared in order to analyze interpretability and scalability.

In the second part of this experiment, we compared the performance of FOLD-SE with XG-Boost for multi-class classification. We compiled seven datasets from Kaggle and the UCI Machine Learning Repository. Specifically, we used the Student Performance Dataset \cite{elkharoua2024students}, the Dry Bean dataset \cite{uci2020drybean}, the Iris dataset \cite{uci2021iris}, the Breast Cancer Wisconsin dataset \cite{uci2019breastcancer}, the Wine dataset \cite{uci2019wine}, the Glass dataset \cite{uci2017glass}, and the Drug dataset \cite{tripathi2020drug}.

To obtain scores and rules for FOLD-SE, the same web API for FOLD-SE was accessed. To ensure compatibility with the API, some minor preprocessing was completed, consisting of changing feature names and adding an extra ID column, as well as converting the tabular data into a standard CSV (Comma-Separated Values) format. This was all performed in Microsoft Excel. The CSV files were then uploaded to the website for model creation and evaluation. Then, rules, accuracy, and F1 scores were all obtained and recorded in a data table.

For comparison with multi-class classification for FOLD-SE, we utilized the XG-Boost algorithm, a leading machine learning library for classification applications. The algorithm consists of building an ensemble of boosted trees in parallel on a dataset to then make uniform predictions \cite{nvidia2019xgboost}. The API for XG-Boost can be accessed programmatically using the XGBClassifier class from the xgboost library \cite{xgboost2024api}.

A Google Colab environment was created to evaluate the accuracy of XGBoost, which was chosen because it contains many popular libraries pre-installed and offers a user-accessible GPU. The model was  created by using a 80/20 train/test split and grid search cross-validation was performed across 5 folds. Then, the optimal hyperparameters for the particular dataset were identified, and the best model was evaluated on the test set to determine its accuracy and F1 score. A weighted F1 score was used to compute the F1 score for the classes in multi-categorical classification.
Using the time module in Python, the start and end times in nanoseconds were tracked. This pipeline was run across all five datasets collected, and the resulting accuracy, F1 score, and elapsed time were tracked and output in the same data table used for FOLD-SE metrics for proper comparison.

The accuracy and F1 scores between the two machine learning algorithms for all five datasets were then analyzed to evaluate their predictive performance. The rules produced by FOLD-SE were also looked at to assess the overall interpretability of the model.  

\section{Results}
We output all results in the tables below. Bolded values emphasize metrics that showcase better performance in the context of model comparison. For sake of readiness, we will split our results into 2 results for each model comparison.
\subsection{Result 1: FOLD-SE vs. FOLD-R++}
\begin{table}[H]
\centering
\resizebox{\textwidth}{!}{%
\begin{tabular}{lcccccc}
\toprule
\textbf{Dataset} & \textbf{FOLD-SE Acc.} & \textbf{FOLD-R++ Acc.} & \textbf{FOLD-SE F1} & \textbf{FOLD-R++ F1} & \textbf{FOLD-SE Time (ms)} & \textbf{FOLD-R++ Time (ms)} \\
\midrule
Adult     & \textbf{84.7\%}  & 84.1\%  & 0.90 & 0.90 & 4780.08 & \textbf{2975.18} \\
Autism    & 84.4\%  & \textbf{93.6\%} & 0.89 & \textbf{0.957} & 82.31   & \textbf{53.27} \\
Ecoli     & 94.12\% & \textbf{95.6\%} & 0.93 & \textbf{0.943} & 37.52   & \textbf{19.37} \\
Heart     & 77.78\% & \textbf{79.6\%} & \textbf{0.79} & 0.776 & 21.22   & \textbf{18.42} \\
Kidney    & \textbf{98.75\%} & 97.5\%  & \textbf{0.99} & 0.978 & \textbf{23.47} & 24.46 \\
Sonar     & \textbf{88.1\%}  & 71.4\%  & \textbf{0.88} & 0.808 & 414.71  & \textbf{255.43} \\
Mushroom  & 99.88\% & \textbf{100\%}  & 1.0  & 1.0  & 629.44  & \textbf{343.44} \\
Raisin    & 82.78\% & \textbf{85\%}   & 0.86 & \textbf{0.868} & 274.41  & \textbf{95.25} \\
Breast    & 92.98\% & \textbf{93.9\%} & \textbf{0.95} & 0.916 & 291.53  & \textbf{221.67} \\
Loan      & \textbf{81.09\%} & 80.8\%  & \textbf{0.88} & 0.878 & 597.76  & 767.77 \\
\bottomrule
\end{tabular}
}
\caption{Comparison of accuracy, F1 score, and elapsed time between FOLD-SE and FOLD-R++. Bolded values indicate better performance.}
\label{tab:foldse_accuracy_time}
\end{table}

\begin{table}[H]
\centering
\resizebox{\textwidth}{!}{%
\begin{tabular}{lcccc}
\toprule
\textbf{Dataset} & \textbf{FOLD-SE Base Rules} & \textbf{FOLD-R++ Base Rules} & \textbf{FOLD-SE Exc. Rules} & \textbf{FOLD-R++ Exc. Rules} \\
\midrule
Adult     & \textbf{2} & 5  & \textbf{0} & 8  \\
Autism    & 8  & 8  & \textbf{6} & 12 \\
Ecoli     & \textbf{1} & 4  & 3  & \textbf{2} \\
Heart     & 1  & 1  & \textbf{0} & 2  \\
Kidney    & \textbf{4} & 5  & 1  & \textbf{0} \\
Sonar     & \textbf{3} & 5  & 3  & \textbf{2} \\
Cars      & \textbf{4} & 13 & \textbf{3} & 5  \\
Mushroom  & \textbf{4} & 5  & 2  & 2  \\
Raisin    & 1  & 1  & 3  & \textbf{0} \\
Breast    & \textbf{1} & 5  & 4  & \textbf{3} \\
Loan      & \textbf{2} & 13 & \textbf{0} & 29 \\
\bottomrule
\end{tabular}
}
\caption{Comparison of rule complexity between FOLD-SE and FOLD-R++. Bolded values indicate fewer rules (better interpretability).}

\label{tab:foldse_rules}
\end{table}

As can be seen in \ref{tab:foldse_accuracy_time} and \ref{tab:foldse_rules}, our first hypothesis that FOLD-SE would perform worse in terms of accuracy and F1 score was invalidated. FOLD-SE had nearly equal or improved accuracy compared to FOLD-R++, all while generating significantly fewer rules. But a downside to FOLD-SE was its increased processing time: 6/8 datasets were processed $\sim$50\% slower using FOLD-SE compared to FOLD-R++. 

We also found that FOLD-SE was able to produce a small set of highly interpretable rules, elucidating the model's decisions. As can be seen below, FOLD-SE produced two clear rules to classify whether an adult's annual income was less than 50k, compared to FOLD-R++, which produced 13. 
\begin{figure}[!htbp]
\centering
\begin{minipage}{0.95\textwidth}
\begin{lstlisting}
label(X,'<=50K') :- not marital-status(X,'Married-civ-spouse'),
                    capital-gain(X,N1), N1=<6849.0.
label(X,'<=50K') :- marital-status(X,'Married-civ-spouse'),
                    capital-gain(X,N1), N1=<5013.0,
                    educational-num(X,N2), N2=<11.0.
\end{lstlisting}
\end{minipage}
\caption{Rules generated by FOLD-SE for the Adult dataset.}
\label{fig:foldse_rules_adult}
\end{figure}

\subsection{Result 2: FOLD-SE vs. XG-Boost}
\begin{table}[!htbp]
\centering
\resizebox{\textwidth}{!}{%
\begin{tabular}{lcccccc}
\toprule
\textbf{Dataset} & \textbf{XGBoost Acc.} & \textbf{FOLD-SE Acc.} & \textbf{XGBoost F1} & \textbf{FOLD-SE F1} & \textbf{XGBoost Time (ms)} & \textbf{FOLD-SE Time (ms)} \\
\midrule
Students Performance & 92.4\%  & \textbf{99.58\%} & 0.92 & \textbf{1.0} & 194045  & \textbf{210.26} \\
Dry Beans            & 93.2\%  & \textbf{99.23\%} & 0.93 & \textbf{1.0} & 1244904 & \textbf{19754.61} \\
Iris                 & 93.3\%  & \textbf{100.0\%} & 0.93 & \textbf{1.0} & 34971   & \textbf{6.31} \\
Breast Cancer        & 96.4\%  & \textbf{100.0\%} & 0.96 & \textbf{1.0} & 200960  & \textbf{703.76} \\
Wine                 & 100.0\% & 100.0\%          & 1.0  & 1.0          & 50270   & \textbf{41.15} \\
Glass                & 72.0\%  & \textbf{100.0\%} & 0.68 & \textbf{1.0} & 127254  & \textbf{97.77} \\
Drug                 & 97.5\%  & \textbf{100.0\%} & 0.98 & \textbf{1.0} & 69036   & \textbf{6.10} \\
\bottomrule
\end{tabular}
}
\caption{Comparison of accuracy, F1 score, and elapsed time between FOLD-SE and XGBoost. Bolded values indicate better performance.}
\label{tab:foldse_xgboost_accuracy_time}
\end{table}

\begin{table}[H]
\centering
\resizebox{\textwidth}{!}{%
\begin{tabular}{lccc}
\toprule
\textbf{Dataset} & \textbf{FOLD-SE Base Rules} & \textbf{FOLD-SE Exc. Rules} & \textbf{Notes} \\
\midrule
Students Performance & 6  & 0 & Compact rule set \\
Dry Beans            & 9  & 2 & Few exceptions \\
Iris                 & 5  & 2 & Smallest dataset, very few rules \\
Breast Cancer        & 9  & 2 & Highly interpretable \\
Wine                 & 5  & 1 & Minimal rules \\
Glass                & 13 & 5 & More rules due to complexity \\
Drug                 & 6  & 1 & Few rules, easy to interpret \\
\bottomrule
\end{tabular}
}
\caption{FOLD-SE rule complexity across multiclass datasets compared with XGBoost (which produces no explicit human-readable rules).}
\label{tab:foldse_xgboost_rules}
\end{table}

In \ref{tab:foldse_xgboost_accuracy_time} and \ref{tab:foldse_xgboost_rules}, we can see that FOLD-SE outperformed XG Boost in accuracy and processing time in all five models we tested, obtaining an average 9.18\% higher accuracy, all while generating a low number of rules for high interpretability and having lower process times.

\section{Discussion}

This research was planned to examine and contrast the effectiveness of FOLD-SE, FOLD-R++, and XGBoost on different classification tasks with multiple datasets. Against our expectations, we found that the accuracy, as measured by the F1 score, for FOLD-SE did not decline when compared to that of FOLD-R++, despite producing fewer rules and improving interpretability. As shown in \ref{fig:foldse_rules_adult}, FOLD-SE was able to achieve improved accuracy while generating only two rules compared to 13 total rules generated by FOLD-R++. This enables the model to be easily interpreted as a clear logical 'flow' that the model takes on can be easily formulated through less rules. Furthermore, this would serve a useful purpose in analyzing any systemic biases existing in the data itself.

This result highlights the importance of FOLD-SE as not just a more efficient tool in creating rules but also an equally good choice in terms of predictability and accuracy.

Our other presumption was that FOLD-SE would show a decrease in accuracy as compared to XGBoost, due to its emphasis on explainability and quicker processing time. But this assumption also proved false when FOLD-SE exceeded the accuracy of XGBoost across all seven evaluated datasets. The fact that FOLD-SE achieved higher accuracy while maintaining lower processing times and greater interpretability demonstrates how more efficient algorithms can be utilized to increase processing time, accuracy, and explainability simultaneously, showcasing new breakthroughs in situating clever logic-based algorithms in ML.

Multiple elements could affect our outcomes. The choice of different datasets, each with distinct correlations and biases, may give an advantage to one model over another. Moreover, the platforms used for conducting experiments (VS Code, Colab, and FOLD-SE online tool) may have introduced variations in terms of processing power and execution time. Future research could gain from having more uniform computational conditions to lower potential biases. 

These findings' outcomes are shown in areas where it's vital for machine learning models to be understandable as well as good at prediction. FOLD-SE can provide precise and explainable models quickly, making it a potential choice for use in industries that must adhere to strict rules and make costly decisions, such as finance and healthcare.

But this study does have its restrictions. Although the datasets used here differ from one another, using a wider range of datasets could facilitate a better understanding of the strengths and weaknesses of each model. 

Further research is needed to see how FOLD-SE and FOLD-R++ perform in practical settings where data may be more unorganized. Additionally, exploring the potential of combining these models with other machine learning techniques could provide a deeper understanding of their utility for tasks involving complex classification.

\section{Conclusion}
Ultimately, this research demonstrates that FOLD-SE is a top-ranking model in tasks involving binary and multi-category classification. It provides an excellent balance between accuracy, explainability, and efficiency. The findings challenge the notion that accuracy and explainability have an inverse relationship, demonstrating that they can be enhanced simultaneously. These discoveries contribute to the growing number of studies advocating for machine learning models to be more transparent in their interpretation, without compromising performance.



\end{document}